# What you need to know about the state-of-the-art computational models of object-vision: A tour through the models.

**Seyed-Mahdi Khaligh-Razavi\***

\*MRC Cognition and Brain Sciences Unit, Cambridge University, Cambridge, UK
 smk49@cam.ac.uk

## Abstract

Models of object vision have been of great interest in computer vision and visual neuroscience. During the last decades, several models have been developed to extract visual features from images for object recognition tasks. Some of these were inspired by the hierarchical structure of primate visual system, and some others were engineered models. The models are varied in several aspects: models that are trained by supervision, models trained without supervision, and models (e.g. feature extractors) that are fully hard-wired and do not need training. Some of the models come with a deep hierarchical structure consisting of several layers, and some others are shallow and come with only one or two layers of processing. More recently, new models have been developed that are not hand-tuned but trained using millions of images, through which they learn how to extract informative task-related features. Here I will survey all these different models and provide the reader with an intuitive, as well as a more detailed, understanding of the underlying computations in each of the models.

## Introduction

The last stage of visual object processing (i.e. inferior temporal) in the ventral pathway produces an invariant representation of objects, even in the face of significant image variations. How is this achieved by the brain, and what are the computational mechanisms underlying this process? To find a computational mechanism that explains the outstanding ability of humans in object recognition, several artificial vision systems have been developed. There are computational frameworks developed in computer vision and computational neuroscience, some of which partly resemble the similarity patterns observed in inferior temporal (IT) cortex of primates. The idea of having an artificial visual system is not very new and has been around since pre-historic times. In classical mythology an artificial visually-guided agent is named that according to the legends has had the role of monitoring boarders of a Mediterranean island to protect it from invaders (Graves, 2012).

Efforts in computational vision has helped designing better artificial object recognition systems, however, much more work is needed to accomplish the task of designing an intelligent machine that sees and processes objects as human does. Object recognition involves scene categorization, image-level annotation, object detection, image parsing; and the system has to deal with all within-class variability (categorization task) while being able to distinguish between different objects of the same class (identification task). Yet today we don't have such a system in place that does all these things at the level of human performance.

The computational models of object-vision can guide our intuition in understanding the brain mechanism in visual information processing. Here, I introduce some of the state-of-the-art object-vision models, and survey the breadth of approaches adopted over the years in attempting to solve the problem of object recognition, and highlight the important role that each of them plays.

The models that I will present in this manuscript are shown in Figure 1. In the first section I will present some of the basic image representations (e.g. silhouette image), and some mathematical transforms applied to images (e.g. radon transform), these are the untrained shallow models shown by the light blue colour (Figure 1). In the next section, state-of-the-art computer vision models (e.g. PHOW) and features extractors (e.g. gist) will be discussed. These models are more to the right side of Figure 1. The last section is about biologically-inspired models of object-vision, and the models discussed in this section (e.g. HMAX, deep convolutional network) are usually hierarchical models that are inspired by the hierarchy of ventral visual pathway of primate cortex.

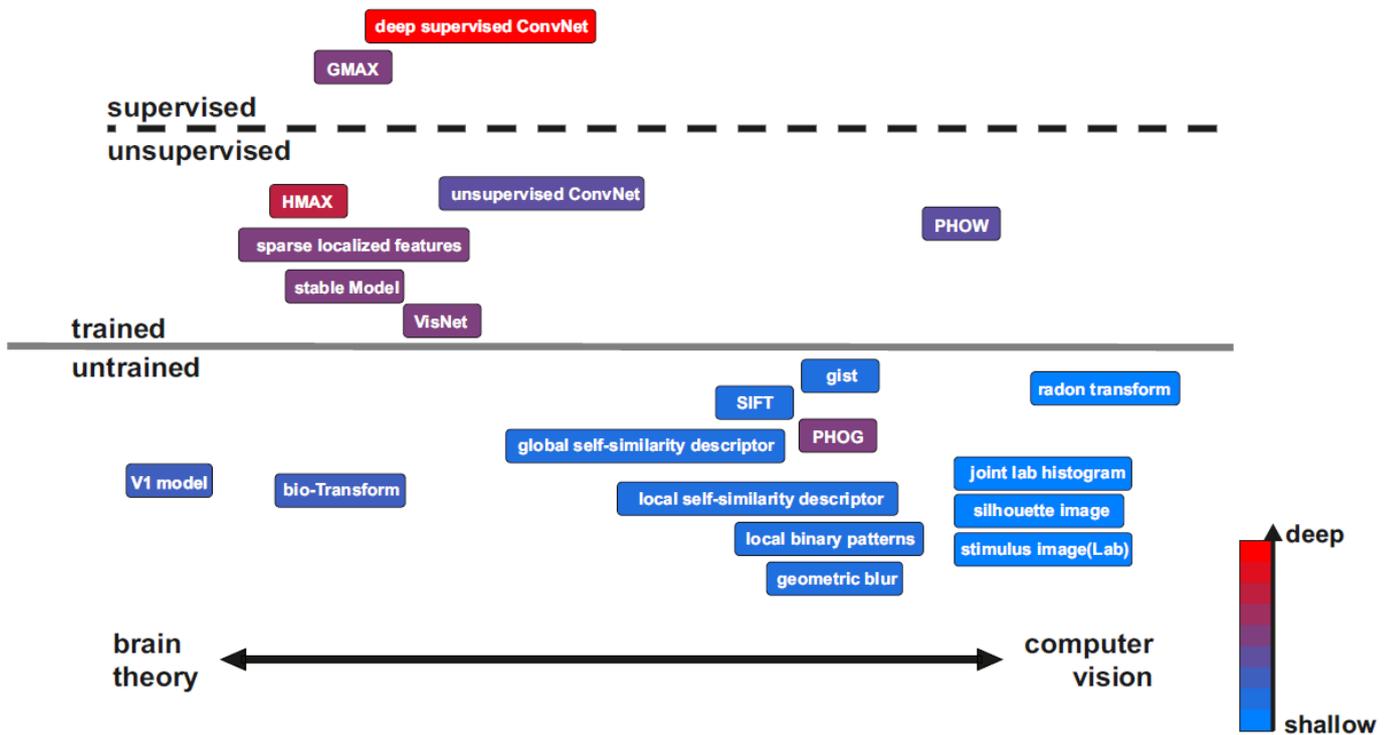

**Figure 1. Descriptive arrangement of the models.** The figure shows all the models discussed in this paper. The models are arranged according to the level of training that they need (untrained, unsupervised trained, supervised trained), their inspiration from biology (biologically-inspired models or engineered models from computer vision), and how deep the models are in terms of the number of processing layers. The depth of the models is color-coded from light blue (shallow) to red (deep). Models to the left hand side are more inspired by the brain as opposed to the models on the right hand side that are engineered models from computer vision.

# Basic image representations

Models in this category provide us with very simple image representations that are close to the original pixel space.

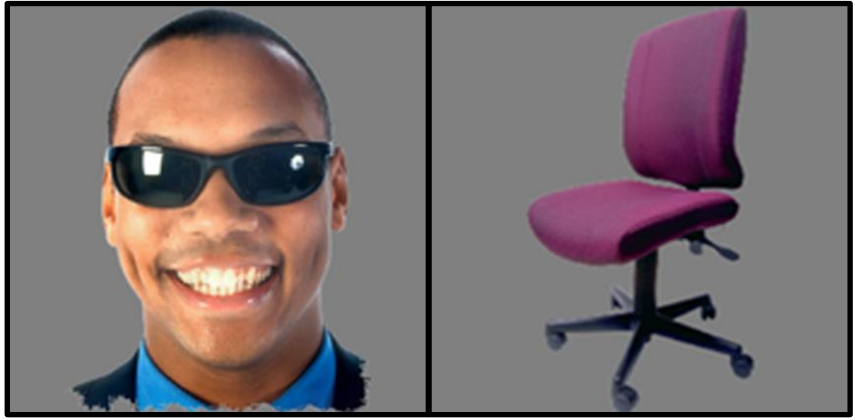

**Figure 2. Two sample images.** Through the paper, I have used these two images to visualize the internal representation of each of the models where applicable.

**Stimulus image (Lab):** Lab color space approximates a linear representation of human perceptual color space, and includes all perceivable colors. A Lab color space has three dimensions: *L* for lightness, *a,* and *b* for color-opponent dimensions. Color-opponent dimensions code the difference between color channels, that is the difference for red versus green, and blue versus yellow. Each Lab image can be obtained by transferring the color image from RGB color space to the Lab color space. Then, the image should be converted to a pixel vector with the length of X*Y*3, where *X*, and *Y* are the width and height of the image.

**Color set (lab joint histogram)**: This is an extension to the Lab color space image representation. First, images (X * Y pixels) are transferred from RGB color space to Lab color space. Then, the three Lab dimensions *(L, a, b)* are divided into 6 bins of equal width. The joint histogram is then computed by counting the number of figure pixels falling into each of the 6x6x6 bins. Finally, the obtained lab joint histogram is converted to a vector with the length of 6x6x6.

**Radon**: The Radon transform is an integral transform. The inverse of Radon has been originally used to reconstruct images from medical CT scans. This transform is not meant to be biologically-inspired; however, it has been proposed as a functional account of the representation of visual stimuli in the lateral occipital complex (Wade and Tyler, 2005).

The transform is applied to each pixel. It first divides each pixel into four subpixels and then projects the value of each subpixel along a radial line oriented at a specific angle. The Radon transform of an image is the sum of the Radon transform of each individual pixel. This gives us a matrix, in which each column corresponds to a set of integrals of the image intensities along parallel lines of a given angle. The Matlab function Radon can be used to compute the Radon transform for each luminance image.

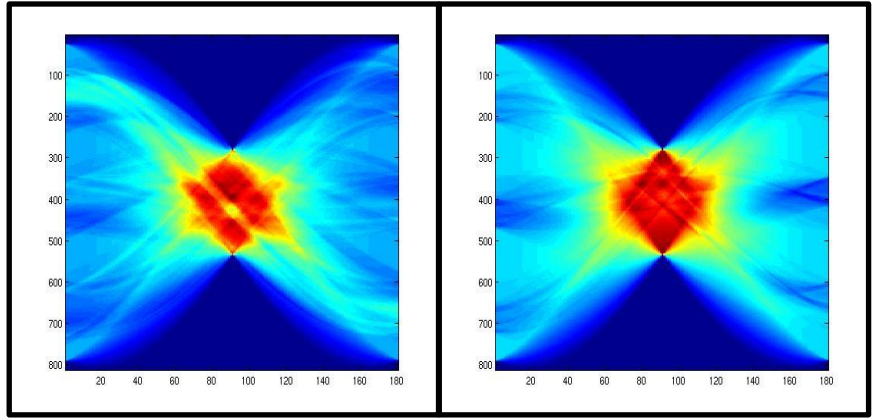

**Figure 3. Radon transform output.** The left panel is the radon transform of the face image, and the right panel is the radon transform of the chair image.

**Silhouette image**: a silhouette of an image, as its name may suggest, is a solid dark shape representation of an image that shows an outline of the objects in the image.

An RGB color image can be converted to a binary silhouette image by setting all background pixels to 0 and all figure pixels to 1.

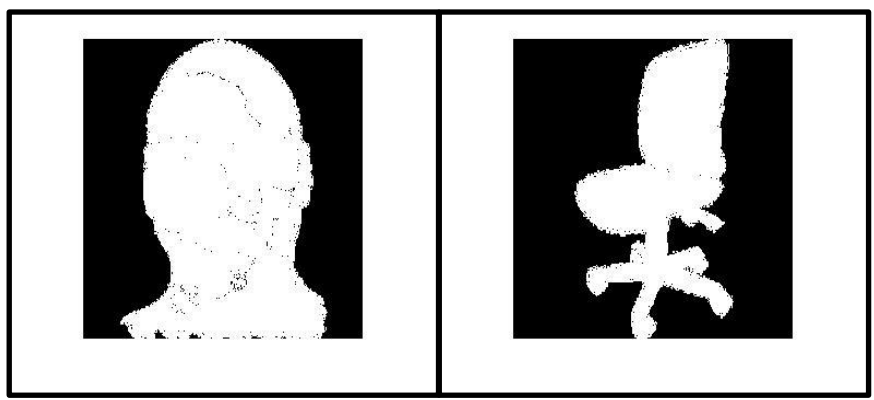

**Figure 4. Silhouette output of the two sample images.**

# Computer vision models

In this section I will explain some of the popular feature extractors and computer vision models. The performance of several local image descriptors, including some of the feature extractors that I will discuss in this section are evaluated by Mikolajczyk and Schmid ( 2005) in the context of matching and recognition of the same scene or object observed under different viewing conditions. Their study is a comprehensive comparison and should provide the reader with further insights regarding the relative ability of computer vision feature descriptors introduced in this section. Furthermore for a recent review about the object recognition models that have been developed in computer vision during the past half-century see (Andreopoulos and Tsotsos, 2013).

**Local Binary Patterns (LBP):** LBP was first described in 1994 (Ojala et al., 1994, 1996). Since then it has been found to be powerful for texture categorization. It is a simple yet efficient texture operator. The underlying idea of LBP is that a 2-dimensional surface can be described by two complementary measures, which are local spatial patterns and gray scale contrast. Two important properties of LBP, which makes it feasible for real-world applications, are: (1) The LBP operation is not computationally costly. This allows LBP to be used in real-time settings. (2) The LBP representation is robust to monotonic gray-scale changes caused by illumination variations.

For a given pixel, LBP descriptor gives binary labels to surrounding pixels by thresholding the difference between the intensity value of the pixel in the center and the surrounding pixels (Ojala et al., 2001, 2002; PietikÃinen, 2010).

The following algorithm explains how an LBP feature vector can be made:

1. Divide the input image into cells, each cell being an X*X window (e.g. 12x12 windows)
2. For each pixel in a cell, compare its value to all pixels in its direct neighborhood. There are 8 surrounding pixels for each pixel (on its left-top, left-middle, left-bottom, right-top, etc.).
3. For those pixels where the value of the center pixel is greater than the neighbor's value, write '1'; otherwise write '0'. This gives an 8-digit binary number for each pixel in a cell. The binary number is usually converted to decimal for convenience.
4. Compute a histogram for each cell, by counting the frequency of the decimal values that are assigned to each pixel in the cell.
5. This is an optional step: you can normalize the histogram.
6. Concatenate the histograms of all cells in a vector.

The method has been also used in several computer vision problems that are not primarily regarded as texture problems, such as face detection and recognition, and motion analysis (Pietikäinen et al., 2011).

There have been also various modifications and extensions of the LBP methodology, for some examples see (Heikkilä et al., 2009; Liao et al., 2009; Mäenpää and Pietikainen, 2004).

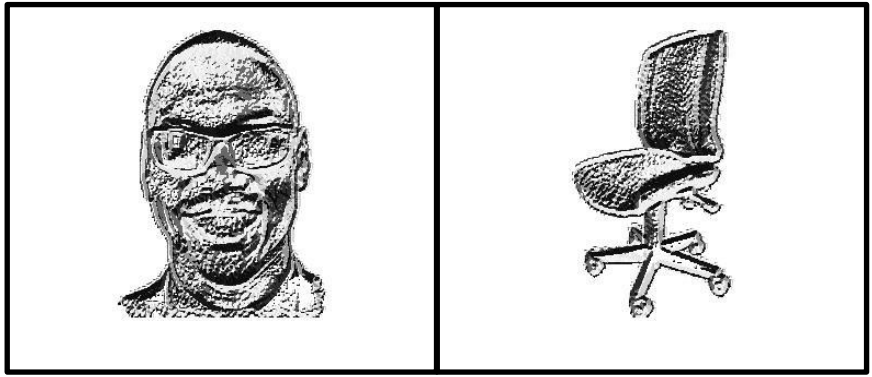

**Figure 5. LBP (local binary pattern) output of the two sample images.**

**Gist:** Human observers can rapidly capture the 'gist' of a scene in a quick feedforward sweep. Therefore a computational model for 'gist' seems a reasonably essential tool for rapid scene classification. Gist has been modelled as average pooling of low-level biologically-inspired features (i.e. gabor-like features) over nonoverlapping subregions arranged on a fixed grid. The term 'spatial envelope' has been also used to refer to this very low dimensional representation of the scene (Oliva and Torralba, 2001). Indeed, gist model bypasses the procedures that are usually applied in scene classification, such as segmentation and processing of individual objects.

The dominant spatial structure of a scene is represented in a set of perceptual dimensions (naturalness, openness, roughness, expansion, ruggedness). The gist model estimates these dimensions using spectral and coarsely localized information.

To calculate the gist features, each image is divided into 16 bins, and then oriented Gabor filters (in 8 orientations) are applied over different scales (4 scales) in each bin. Finally, the average filter energy in each bin is calculated (Oliva and Torralba, 2001, 2006).

There has been some extensions to the gist model: for example, Han and Liu (Han and Liu, 2010) suggested a hierarchical gist model, which was later followed by introducing a task-oriented gist model (Han and Liu, 2013).

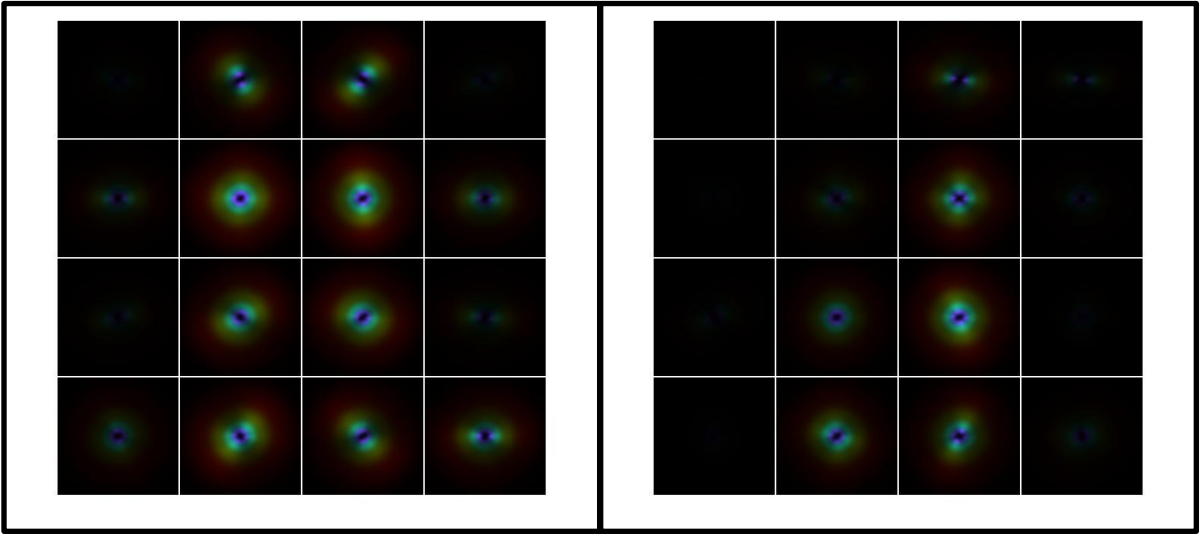

**Figure 6. Gist model output.** The plot is color coded, with one color per scale.(4 scales, 8 orientations ). The left panel is the gist output for the face image, and the right panel is the gist output for the chair image.

**Geometric Blur (GB)**: Geometric blur descriptor comes from a core idea stating that object categorization is fundamentally a problem of deformable shape matching, where related but not identical shapes can be deformed into alignment using simple co-ordinate transformations. The descriptor can be simply defined as an average over geometric transformations of a signal. This can be a useful operation where two signals are being compared and some distortions are expected to be seen in the signals. The descriptor selects some interest points and uses local image properties of the selected interest points. To incorporate some global properties of an image, the relative locations of the interest points are also included in the encoded features.

Geometric blur features can be calculated by selecting a set of uniformly distributed points on an image (as the interest points), and then applying spatially varying blur around the feature points (Belongie et al., 2002; Berg et al., 2005; Zhang et al., 2006). The GB descriptor, around each interest point, regularly samples some pixel values that are then blurred over space with increasing blur for pixels that are further from the interest point. This allows for accumulating more detailed information around the interest points and some coarse context from the surrounding region. This is similar to the decrease in spatial resolution away from the retina's focal point in early vision.

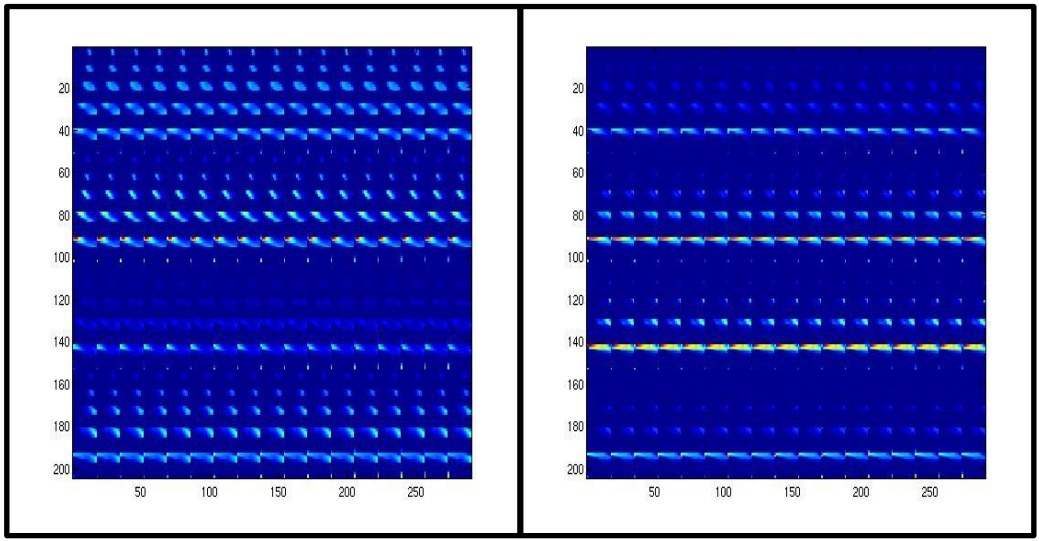

**Figure 7. Geometric blur output.** Left for the face image and right for the chair.

**Scale Invariant Feature Transform:** This is an algorithm suggested by David Low in 1991(Lowe, 1999, 2004) to detect and describe local features in images. SIFT features are suggested to be invariant to changes in scale, orientation, and partially invariant to changes in illumination and affine distortion. The features can be extracted from some interest points or sampled densely from all over the image. It has been shown that dense sampling provides comparable or even better performances than interest points (Yap et al., 2010). One explanation for this could be that having a larger set of local image descriptors that are computed over a dense grid can potentially be more informative about the scene content than the local descriptors computed sparsely over some interest points.

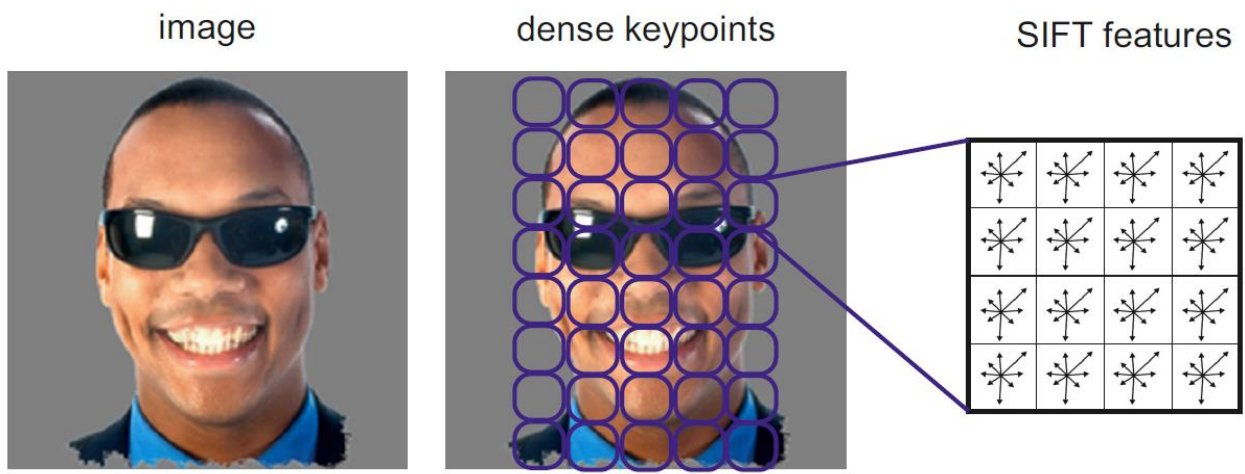

**Figure 8. Extracting dense SIFT features**. This figure depicts how SIFT features are calculated for a sample input image.

To calculate SIFT features, *PxP* pixel patches can be sampled uniformly on a regular grid. Then the histogram of local gradient directions is computed at all the *PxP* grid points around the interest point. Then, all the descriptors are concatenated in a vector as the SIFT representation of the image. Figure 8 shows how dense SIFT features are calculated.

Different works has been done to extend the SIFT descriptor from grey-level to color images. For example, SIFT features can be computed over all three channels in the HSV color space, resulting in a feature vector three times larger than the grey-SIFT(Bosch et al., 2006). Another way would be to concatenate the SIFT descriptor with either weighted hue or opponent angle histograms (Weijer and Schmid, 2006).

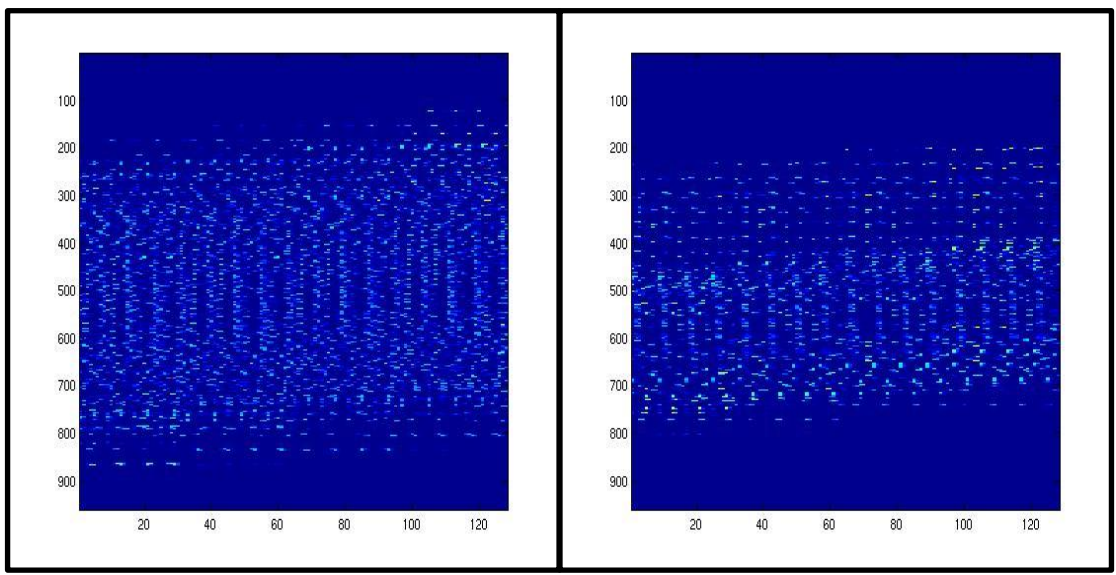

**Figure 9. Dense SIFT output**. There were 31*31 = 961 SIFT patches of size 16x16 over each image. SIFT patches had 8 orientations, and number of bins for histograms were 4. The SIFT output for the face image is shown in the left panel, and the chair is shown in the right panel.

**Pyramid Histogram of Visual Words (PHOW):** PHOW is an extension to the bag-of-words (BoW) model, in which image features are treated as words. The main idea has been originally suggested for document classification in text understanding, and has been later applied to image recognition in computer vision. A bag of words (image features) is a sparse vector of the frequency of the words (image features) repeated in a text document (image). The main problem with the bag of visual words model is that the spatial information of image features are no longer available in the model representation. In BoW we know that a particular feature exist in the image, and we know how frequently, but we can't say where in the image. Pyramid histogram of visual words has been suggested to address this problem. This approach works by dividing the image into increasingly fine sub-regions, which are called pyramids. The histogram of visual words is then computed in each local sub-region.

In order to calculate PHOW features, dense SIFT descriptors can be calculated for each of the training images and then be quantized using k-means clustering to form a visual vocabulary. A spatial pyramid of three levels (could be more or less) should then be created. The histogram of SIFT visual words that is calculated for each bin forms PHOW features. Despite the simplicity of the model it has shown promising results on large-scale datasets (Lazebnik et al., 2006).

**Pyramid Histogram of Gradients (PHOG):** Histogram of oriented gradients are feature descriptors used for object detection. It was first introduced by Navneet Dalal and Bill Triggs, researchers for the French National Institute for Research in Computer Science and Control (INRIA), 2005 CVPR paper (Dalal and Triggs, 2005). The technique works by counting the occurrence of gradient orientation computed on a dense grid of uniformly spaced cells on an image. The idea behind this algorithm is that the local appearance of objects in an image can be described using the distribution of edge directions. The HOG descriptor has been shown to be in particular useful for pedestrian detection (Dalal and Triggs, 2005; Zhu et al., 2006).

Pyramid histogram of gradients (PHOG) is an extension to HOG features. Extending HOG to PHOG is by analogy very similar to the extension of HOW (histogram of visual words) to PHOW. In PHOG, the spatial layout of the image is preserved by dividing the image into sub-regions at multiple resolutions, and applying the HOG descriptor in each sub-region.

To computer the PHOG features, the canny edge detector is usually applied on grayscale images, then a spatial pyramid is created with four levels (Bosch et al., 2007). The histogram of orientation gradients is then calculated for all bins in each level. All histograms are then concatenated to create the PHOG representation of the input image.

**Local Self-Similarity descriptor (ssim):** This is a descriptor that is not directly based on the image appearance; instead, it is based on the correlation surface of local self similarities.

The underlying assumption in most of the feature descriptors is the existence of a usual photometric property (i.e. pixels colors, intensities, edges, gradients or other filter responses) which is shared between images of the same category, and therefore this can be extracted by the feature extractor and compared across images. This assumption seems restrictive in occasions where images have an object in common but there is no obvious shared image property. For example, when the consisting elements of an object in one image is totally different from another image (e.g. a heart made from candles on a birthday cake, and a heart made by chocolate powder on the top of a cup of coffee). However, the two images have repeated local intensity patterns in nearby image locations. Also, their relative geometric layout is similar locally. Local self-similarity patterns are suggested to solve this problem; instead of using the image appearance directly, this feature descriptor generates a correlation surface of local self-similarities from intensity patterns across some interest points on the image (Shechtman and Irani, 2007).

For computing local self-similarity features at a specific point on the image, say *p*, a local internal correlation surface can be created around *p* by correlating the image patch centered at *p* to its immediate neighbors (Chatfield et al., 2009; Shechtman and Irani, 2007). The ssim features can also be computed uniformly at every 'n' pixels in both X and Y directions.

Junejo et al. have extended the idea of self-similarity patterns from spatial domain to temporal domain. Using temporal self-similarities, they perform human action recognition in video (Junejo et al., 2011). Self-similarity features have also been used for creating visual words in the bag-of-visual words framework(Chatfield et al., 2009).

Overall, self-similarity descriptor is useful as a basis for a fast and scalable object recognition or object detection system. The system is robust to non-rigid deformations of objects and allows searching for objects that are similar in shape despite large changes in texture, colors and pose.

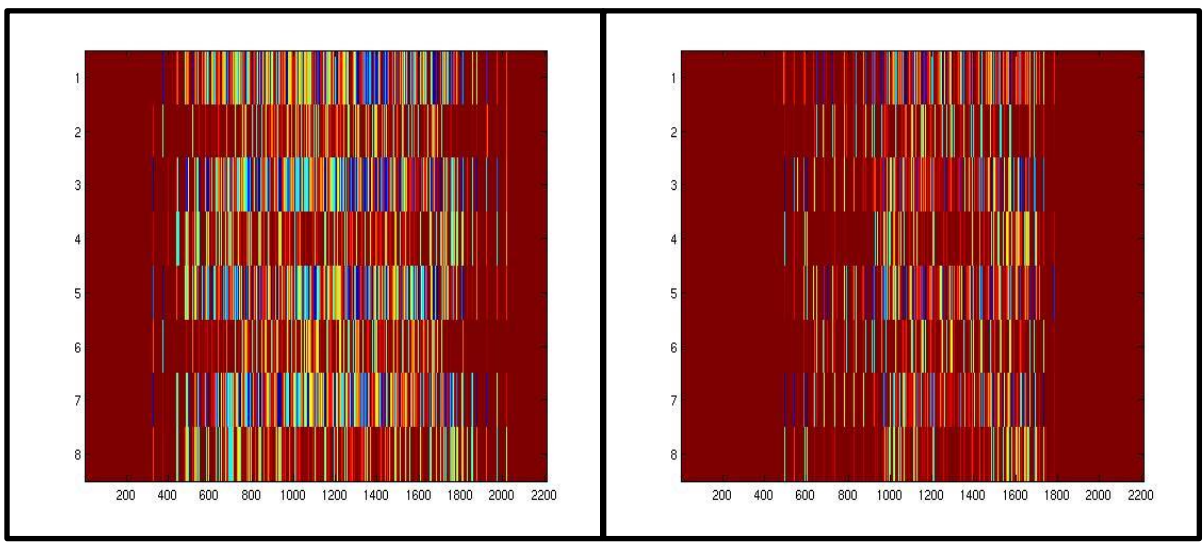

**Figure 10. Local self-similarity output**. The left panel shows ssim output for the face image, and the right panel for the chair.

**Global Self Similarity descriptor (gssim):** This descriptor is an extension to the local self-similarity descriptor mentioned above. The local self-similarity captures self-similarities within relatively small (40x40 pixels) regions. Gssim, however, uses self-similarity globally to capture the spatial arrangements of self-similarity and long range similarities within the entire image (Deselaers and Ferrari, 2010). The authors suggest that in order to take full advantage of self-similarity features they should be computed globally rather than locally. This however makes the gssim features computationally more expensive. To empirically compare local ssim versus global ssim features, both have been tested on Pascal VOC 2007 and on ETHZ Shape Classes dataset. The results suggest that the global ssim outperform the locale features in classification and detection.

Similar to local ssim, global ssim features can also be used as complementary features in conventional image descriptors, such as BoW.

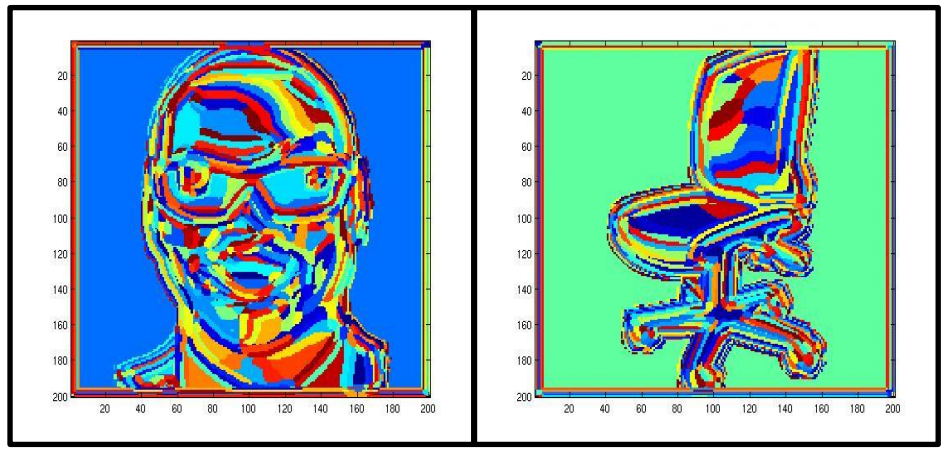

**Figure 11. Global self-similarity output**. This is the visualization of the prototype assignment map for the global self-similarity features. The similarity is color-coded; similar colors mean similar shapes.

# Biologically-inspired object-vision models

**Convolutional networks**: Convolutional neural networks (CNN) were inspired by the hierarchical structure of primate vision and were first introduce by Fukushima 1980 (Fukushima, 1980). The model was called Neocognitron. CNN was later improved by Yann Lecun and his colleagues in 1998 (Lecun et al., 1998) where they used the network for document recognition. From that time onwards it has been under development by several groups (Behnke, 2003; Simard et al., 2003). In recent years by taking advantage of GPU power (Cireşan et al., 2011) people were able to make the model deeper (adding more layers to the model) and train it with more training samples. This has resulted in getting impressive performances in several machine learning tasks (Schmidhuber, 2012).

A major source of difficulty in computer vision has been developing hand-engineered features, like sophisticated feature extractors, to identify higher-level patterns that are optimal for machine vision tasks, such as object recognition. However, convolutional neural networks aim to solve this problem by learning higher-level representations automatically from data.

Convolutional networks are a family of hierarchical models of several stages of feature extraction, each of which is formed by random convolutional filters and subsampling layers (Jarrett et al., 2009). Convolutional layers scan the input image inside their receptive field. Receptive Fields (RFs) of convolutional layers get their input from various places on the input image, and RFs with identical weights make a unit. The outputs of each unit make a feature map. Convolutional layers are then followed by subsampling layers that perform a local averaging and subsampling, which make the feature maps invariant to small shifts (Bengio et al., 1995).

Convolutional neural networks come with many flavors: deep convolutional neural networks (Bengio and Courville, 2013), supervised convolutional networks (Krizhevsky, A., Sutskever, I. and Hinton, G. E., 2012), unsupervised convolutional networks (Lee et al., 2009), and recursive deep neural networks (Pinheiro and Collobert, 2014). During the last decade they have been applied to tackle all sort of different problems in machine learning (Bengio et al., 1995).

In general convolutional networks exploit the three following tricks to yield some degree of invariance: 1) the idea of having local receptive fields pooling from the layer below, 2) having shared weights or weight replication, 3) and spatial or temporal subsampling.

In computer vision, a deep neural network works by feeding an input image through a hierarchy of modules where we want to extract higher and higher levels of features (e.g. edges, then mid-level features, and eventually objects), followed at the end by a classifier that makes a decision. The idea behind the deep learners is to learn these features by training via presenting several images to the system instead of engineering the features by hand. The algorithms scale well with data and computing power (GPU), meaning that with larger datasets and a greater computational power we can get better and better results.

In recent years, a deep supervised convolutional neural network, trained with 1.2 million labelled images from imageNet (1000 category labels) has been proved to be very successful, and has achieved top-1 and top-5 error rates on the imageNet data that is significantly better than previous state-of-the art results on this dataset (Krizhevsky, A., Sutskever, I. and Hinton, G. E., 2012). The network has 8 layers: 5 convolutional layers, followed by 3 fully connected layers. The output of the last layer is a distribution over the 1000 class labels, which is the result of applying a 1000-way softmax on the output of the penultimate layer. The model has 60 million parameters and 650,000 neurons. The parameters are learnt with stochastic gradient descent. The internal representations for different layers of this neural network are shown in Figures 12 to 19.

In the supervised version of the deep CNN, the features in different layers of the model are learnt according to some top level objectives (e.g. assigning labels to images according to their category). We have a bunch of image labels that we want to fit and then we back propagate our gradients through the network to fit all these parameters in the middle to find not only the best classifier but also the best set of features that works well for the classification task.

Deep learning networks have also the ability to take advantage of unlabeled data to learn representations before the task. The networks can be trained with unlabeled data (unsupervised learning) by setting out objectives other than classification. For example the following objectives are desirable for model representations: 1) distributed representation, because ideally we want 'k' features to be able to represent more than 'k' type of patterns; 2) invariant representation, because this makes the model output robust to local changes of input. 3) disentangling factors: we would like to have separate concepts (e.g. color, edge orientation) in separate features (Bengio et al., 2012). We can then minimize an unsupervised training loss based on generic priors about the characteristics of good features.

## 1st conv layer output

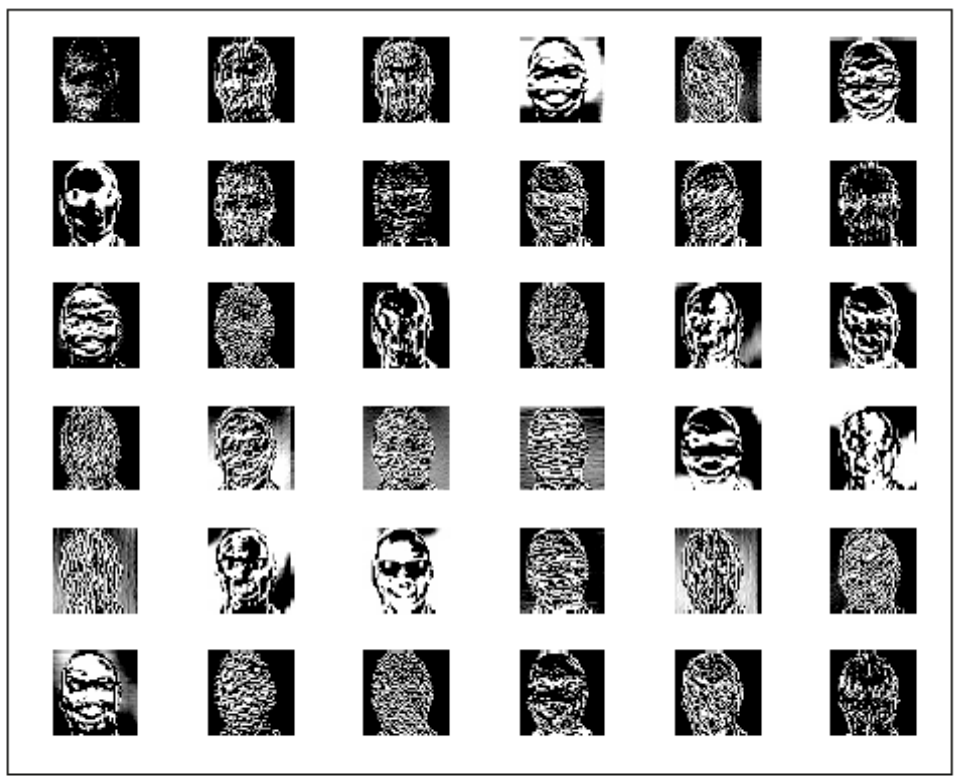

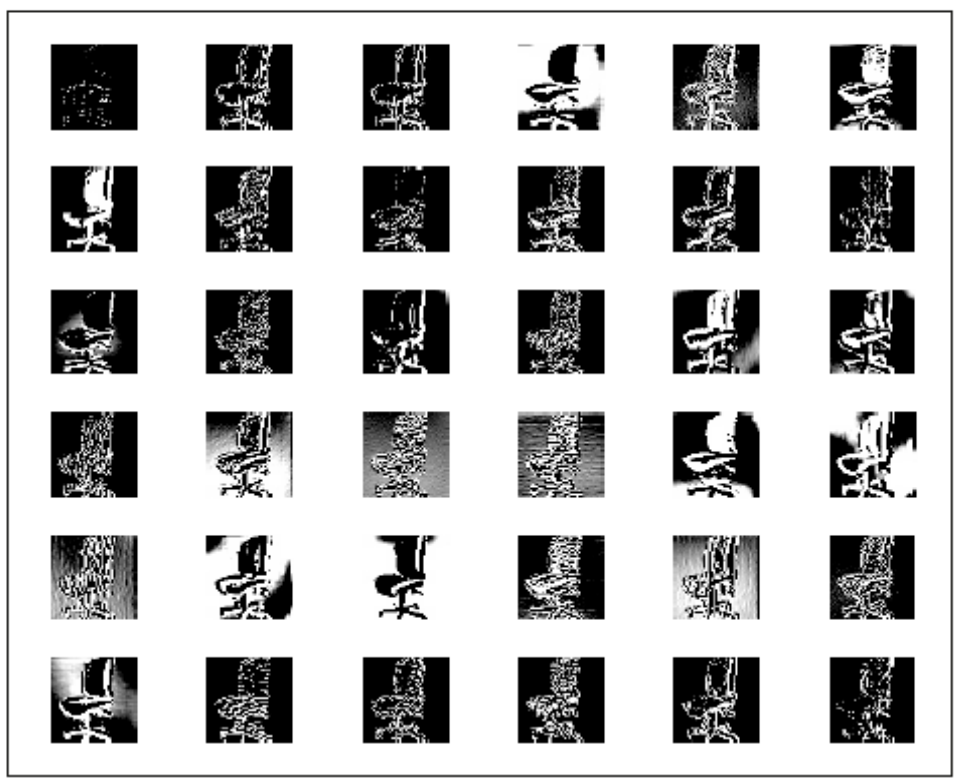

**Figure 12. Deep convNet 1$^{st}$ layer output for (A) the face, and (B) the chair.** These are responses of the first 36 filters out of 96.

## 2nd conv layer output

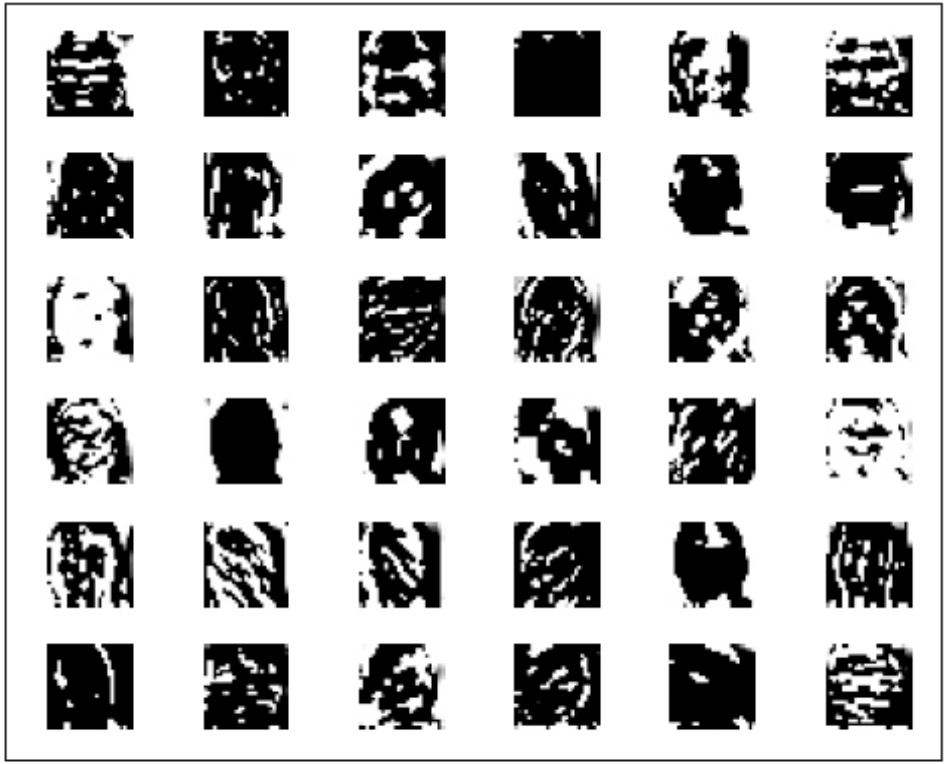

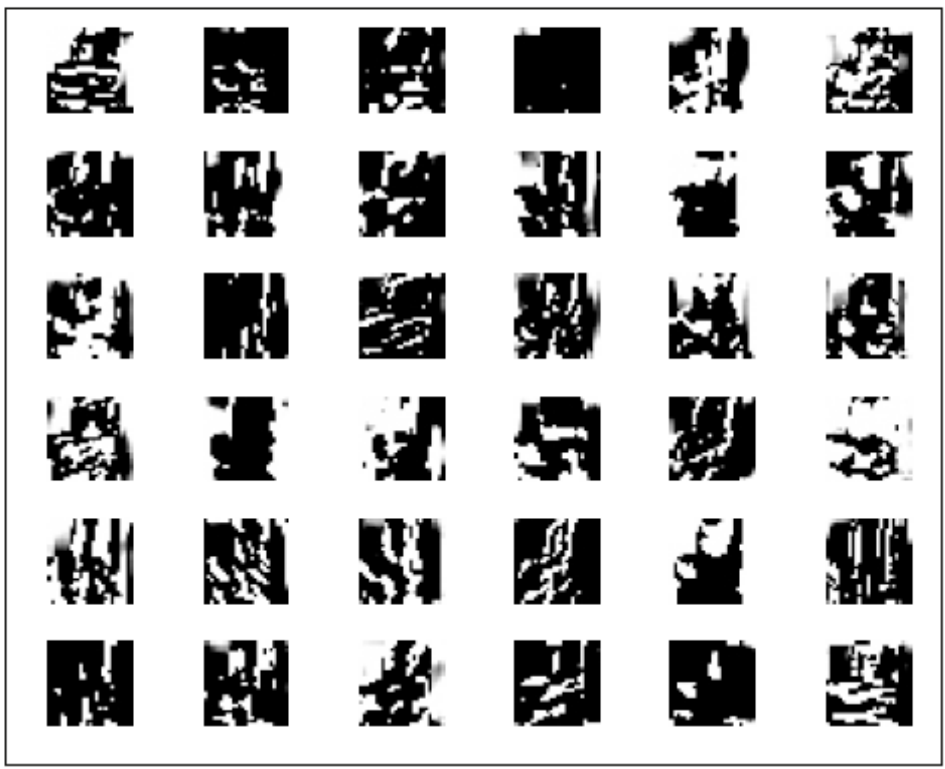

**Figure 13. Deep convNet 2nd layer output for (A) the face, and (B) the chair.** These are responses of the first 36 filters out of 256.

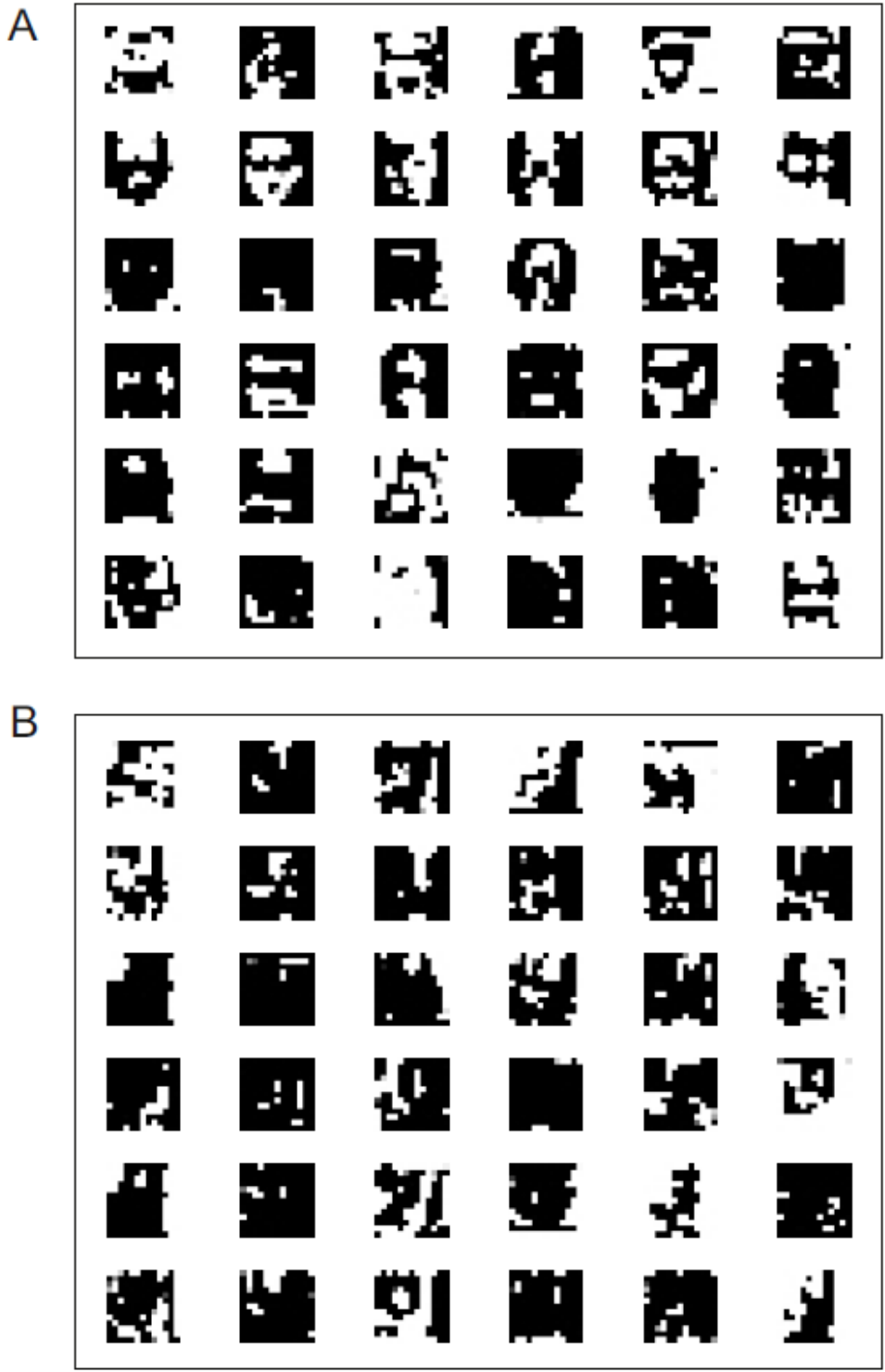

**Figure 14. Deep convNet 3rd layer output for (A) the face, and (B) the chair.** These are responses of the first 36 filters out of 384.

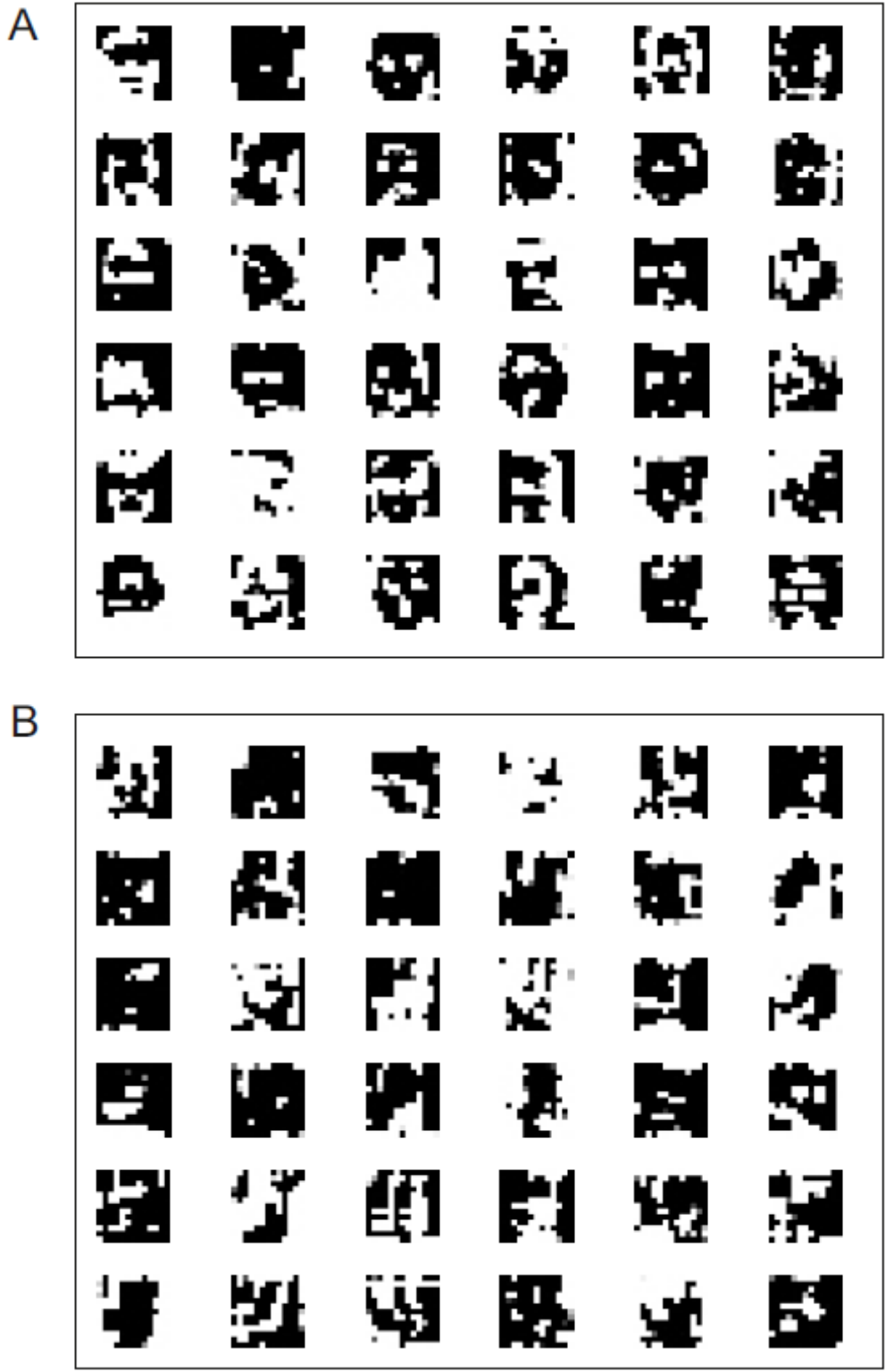

**Figure 15. Deep convNet 4$^{th}$ layer output for (A) the face, and (B) the chair.** These are responses of the first 36 filters out of 384.

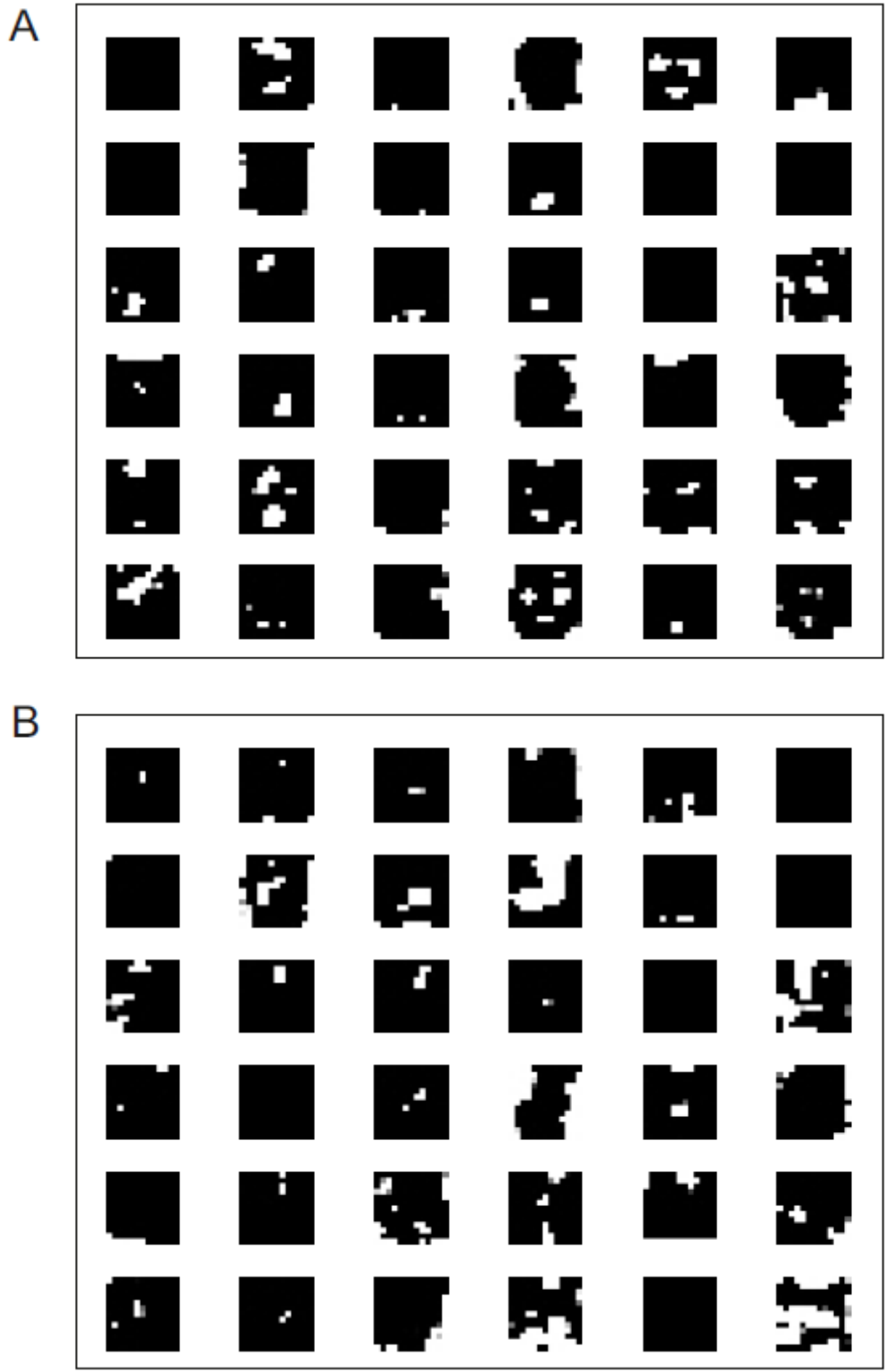

**Figure 16. Deep convNet 5$^{th}$ layer output for (A) the face, and (B) the chair.** These are responses of the first 36 filters out of 256.

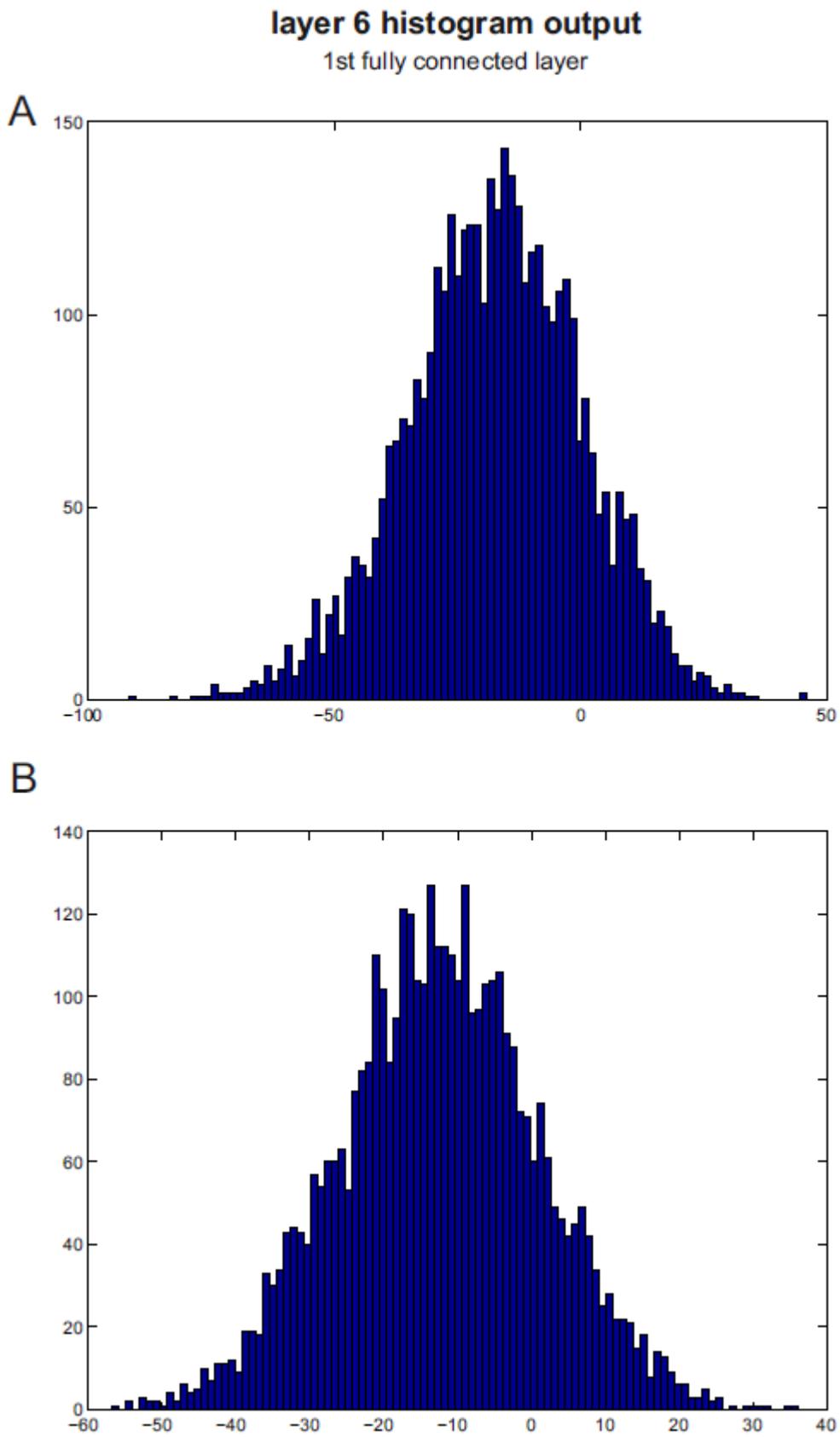

**Figure 17.** Deep convNet histogram of the 6[th] layer output for (A) the face, and (B) the chair.

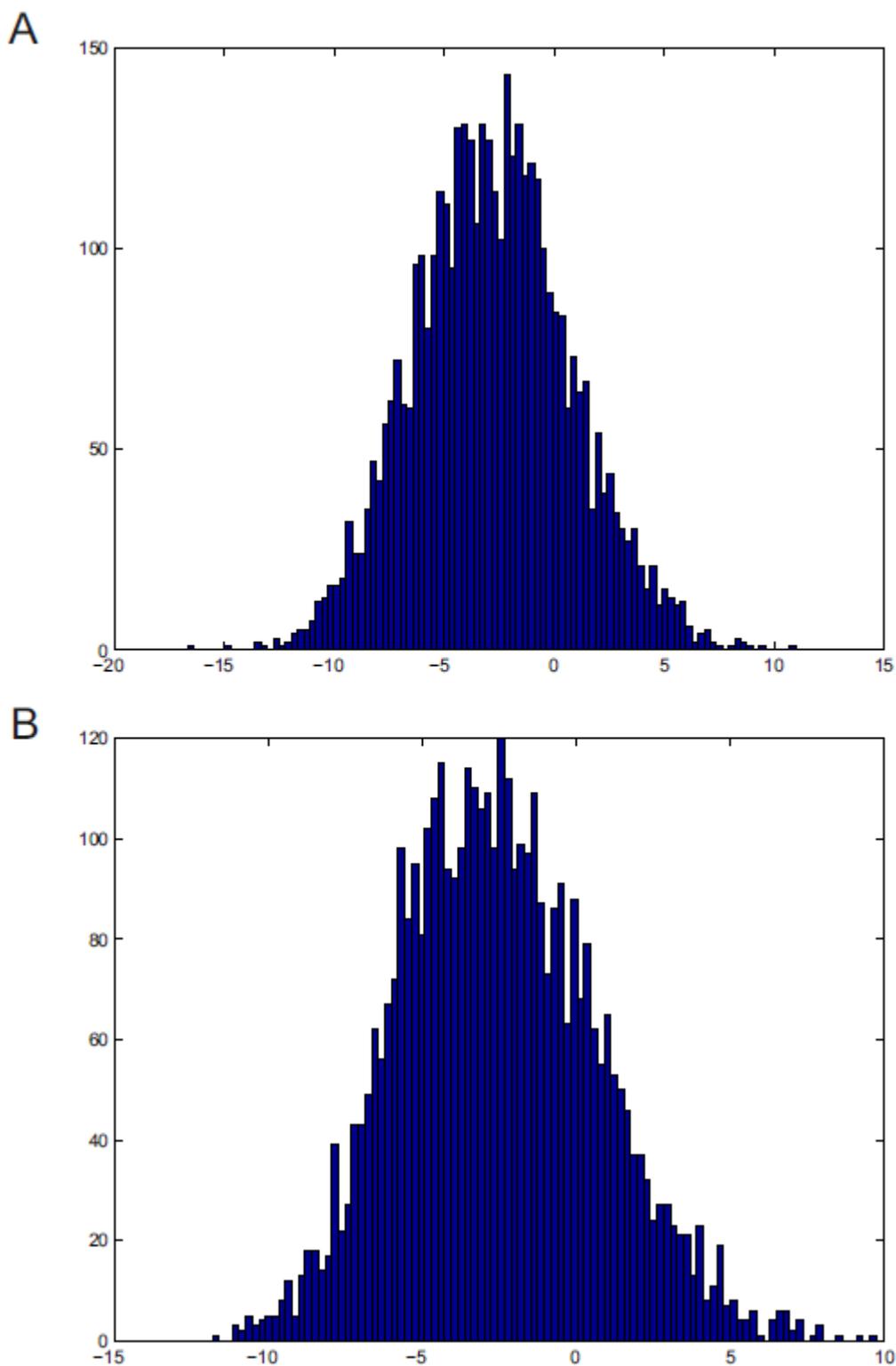

**Figure 18. Deep convNet histogram of the 7$^{th}$ layer output for (A) the face, and (B) the chair.**

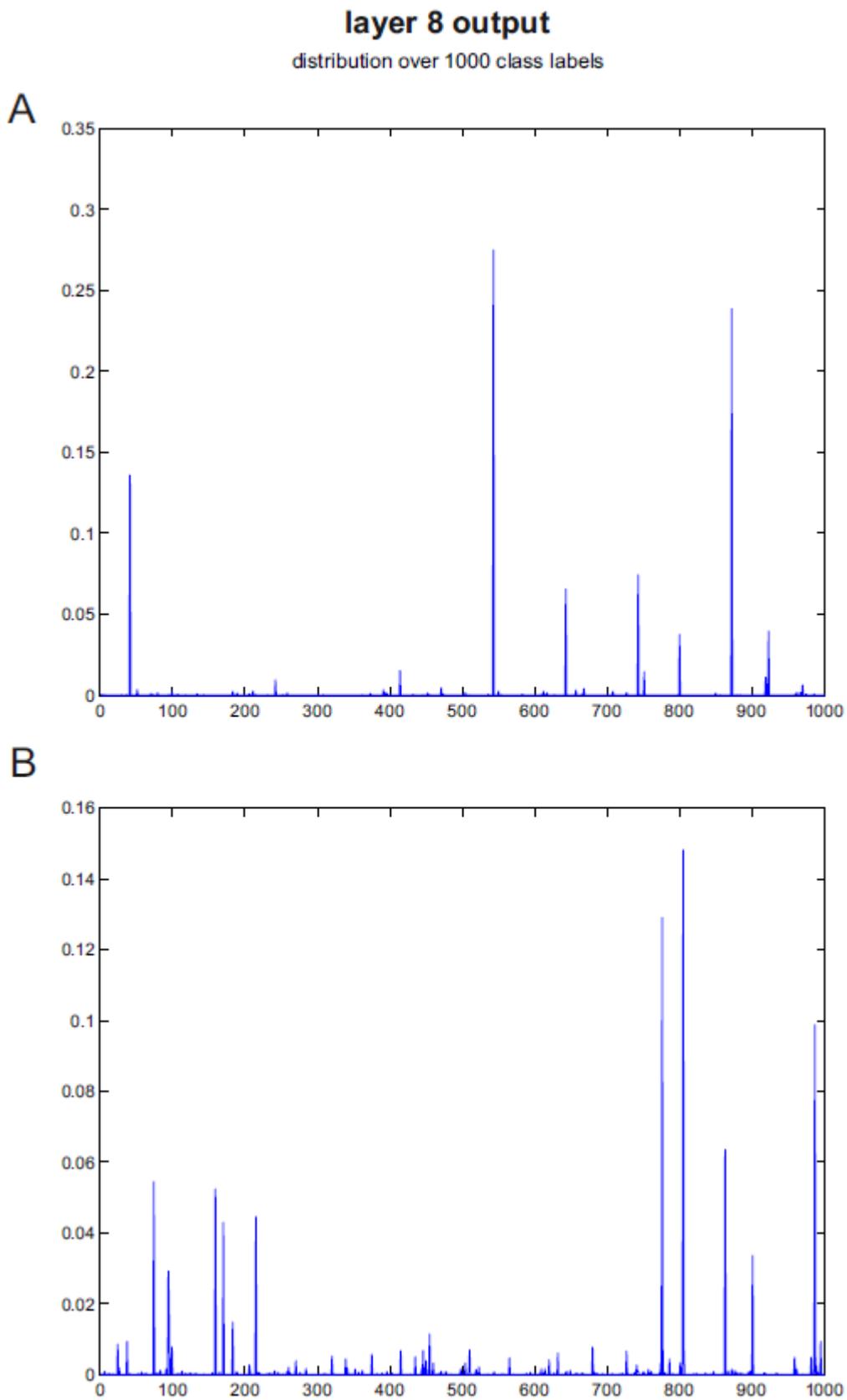

**Figure 19. Deep convNet 8<sup>th</sup> layer output for (A) the face, and (B) the chair**. The output of this layer is a distribution over the 1000 classes that the model is trained for.

**Biological Transform (BT)**: This is a hierarchical model consisting of a set of transforms that make an invariant representation of the input image in a neurally plausible way. The representation is designed to be invariant to position, size, and rotation. The transform is based on local spatial frequency analysis of oriented segments and logarithmic mapping. The model has two stages in each of which the mentioned transform is applied once.

The model is similar to Fourier-Mellin transform in computer vision, except that instead of using a 2D Fourier transform, orientation sensitive cells are used to produce a form of local spatial frequency analysis, which is called interval detection. The orientation-sensitive cells are similar to those in V1, so the model is supposed to mimic the early visual processing stages of primate vision.

In each of the two stages of the model there is an edge detector followed by an interval detector (Sountsov et al., 2011). The edge detector is consisted of a bar edge filter and a box filter. For a given interval *l* and angle *θ*, the interval detector finds edges that have angle *θ* and are separated by an interval *l*. In the first stage, for any given *θ* and *l*, all pixels of the filtered image are summed and then normalized by the squared sum of the input. They are then rectified by the Heaviside function. The second stage is the same as the first stage, except that in the first stage *θ* was changing between 0-180 ° and *l* between 100-700 pixels and the input to the first stage had not a periodic boundary condition on the *θ* axis (repeating the right-hand side of the image to the left of the image and vice versa); but in the second stage the input, which is the output of the first stage, is given a periodic boundary condition on the *θ* axis, and *l* is changing between 15-85 pixels.

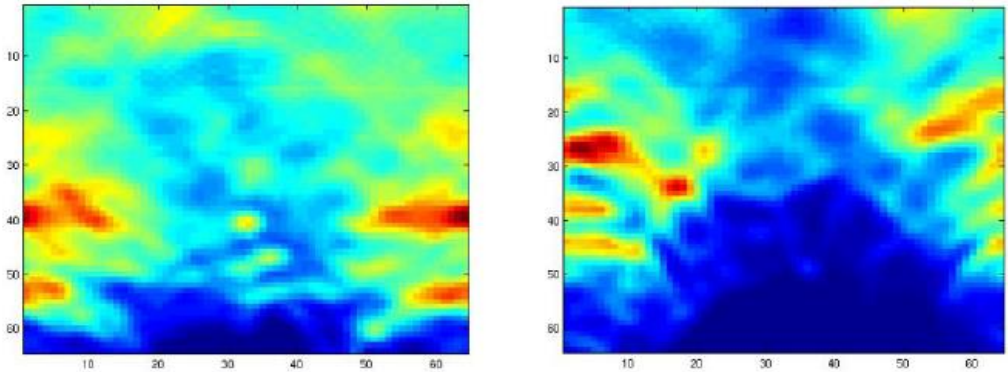

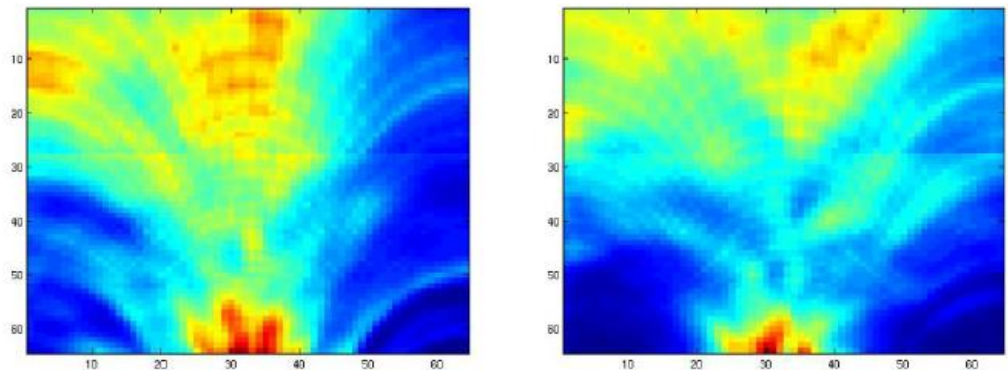

**Figure 20. Bio-Transform output.** The output of the first layer and the second layer of BioTransform is shown (the left panels show the face image; the right panels show the chair image).

**VisNet:** VisNet is a hierarchical model of ventral visual pathway for invariant object recognition that has four successive layers of self organizing maps. Neurons which are higher in the hierarchy have larger receptive fields. Each layer in the model corresponds to a specific area of the primate ventral visual pathway in terms of the size of its receptive fields (Tromans et al., 2011; Wallis and Rolls, 1997). The model can be trained with variations of Hebbian learning rule: trace learning rule (Stringer et al., 2007) and continues transformation learning rule (Stringer et al., 2006) are two usual ways of training the model.

The model is designed based on the following hypotheses to investigate computational aspects of the ventral stream visual processing:

- A hierarchical architecture of competitive networks with mutual inhibition within each layer. Higher-order spatial properties of input stimuli are represented in the network by allowing the neurons to learn combination of features or inputs that occur in a given spatial arrangement.

- Localized population of cells in early layers are connected to the cells in the next layers allowing for an increasing receptive-field size through the hierarchy of visual processing areas.
- A modified Hebbian-like learning rule is used that traces the temporal activity of cells. This enables the neurons to learn transform invariances (Foldiak 1991).

With this model, taking advantage of trace learning rule, it has been shown that it can produce neurons that generate view invariant and translation invariant representations. In investigations of translation invariance described by Wallis and Rolls (1997), the stimuli were simple stimuli such as, T, +. The model has been later extended and applied to more complicated stimuli by Stringer and his colleagues (Stringer et al., 2007).

**V1 model**: Area V1 is the first stage of cortical processing of visual information in the primate and is the gateway of subsequent processing stages. The V1 model is a basic representation that is inspired by known properties of V1 simple and complex cells.

In this model, the population of simple and complex cells are modelled and fed by the luminance images as inputs. Gabor filters of 4 different orientations ($0^o$, $90^o$, $-45^o$, and $45^o$) and 12 sizes (7-29 pixels) are usually used as simple cell receptive fields. Then, the receptive fields of complex cells are modelled by performing the MAX operation on the neighboring simple cells with similar orientations. The MAX operation consists in selecting the strongest (maximum) input to determine the output. This generates a representation invariant to the precise location of the stimulus parts.

The outputs of all simple and complex cells are concatenated in a vector as the V1 representational pattern of each image.

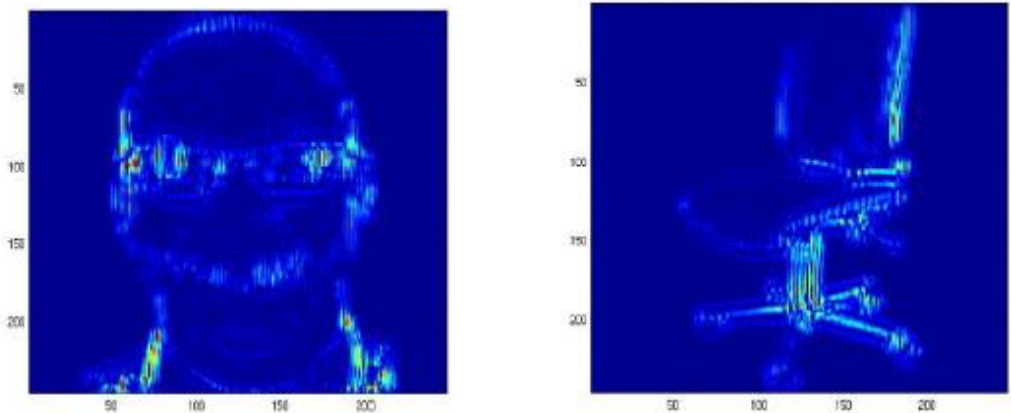

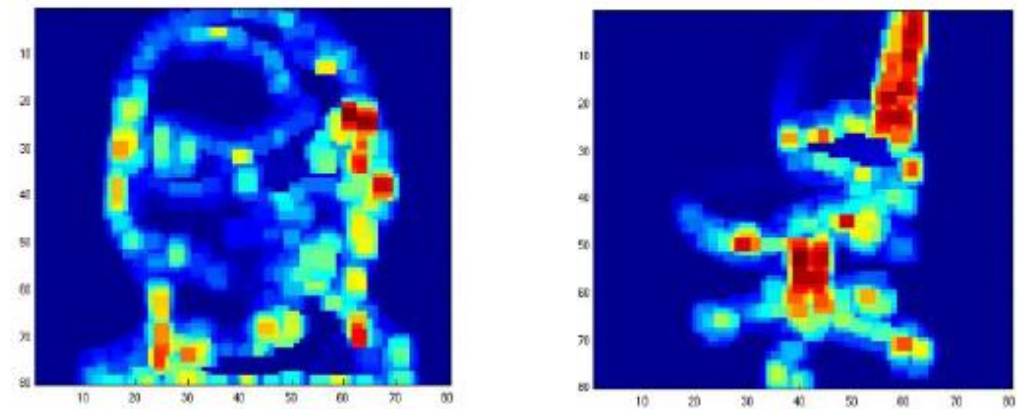

**Figure 21 .V1 model output**

**Gabor wavelet pyramid model (GWP):** The Gabor wavelet pyramid model was used in Kay et al. (2008) to predict responses of voxels in early visual areas in humans.

Gabor wavelets are directly related to Gabor filters, since they can be designed for different scales and rotations. The aim of GWP is to model early stages of visual information processing. Indeed it has been shown that 2D Gabor filters can provide a good fit to the receptive field weight functions found in simple cells of cat's striate cortex (Jones and Palmer, 1987).

GWP model represents each image by a set of Gabor wavelets differing in size, position, orientation, spatial frequency and phase. Kay et al. (2008) used a set of Gabor wavelets of six spatial frequencies, eight orientations and two phases (quadrature pair) at a regular grid of positions over the image to characterize responses of voxels in early visual areas. To control gain differences across wavelets at different spatial scales, the gain of each wavelet is scaled such that the response of that wavelet to an optimal full-contrast sinusoidal grating is equal to 1. The response of each quadrature pair of wavelets is then combined to reflect the contrast energy of that wavelet pair. The outputs of all wavelet pairs are concatenated to have a vector of GWP features for each image.

**HMAX model, its predecessors and successors:** The first qualitatively described model of simple and complex cells in the primary visual cortex was introduced by Hubel & Wiesel (Hubel and Wiesel, 1959, 1968). They described a hierarchical model started by radially symmetric cells responding to the light spots, like center-surround cells in the lateral geniculate nucleus. It was followed by simple cells that respond to oriented bars within their receptive field in a particular position and phase. The next two stages of the hierarchy were complex cells and hypercomplex cells. Complex cells were invariant to position and phase of the bar, and hypercomplex cells were selective for the length of the bar. Since this pioneering work, several hierarchical models of visual object recognition have been developed, among which the model proposed by Riesenhuber and Poggio (1999) is probably the most popular one. The model was later called HMAX ("Hierarchical Model and X") by Mike Tarr (*Nature Neuroscience*, 1999) in his News & Views on the paper. In its simplest architecture, the HMAX model consists of a hierarchy of four layers of computational units (S1, C1, S2 and C2) in order to, firstly, increase specificity and ,secondly, invariance along the hierarchy. The S units combine their inputs and perform Gaussian-like operation to increase object selectivity and build more complex features from simple ones, while C units perform a nonlinear MAX pooling operation over the units tuned to the same feature but at different positions and scales to impose translation and scale invariance.

  An extended version of The HMAX model has been later suggested by Serre et al. (Serre et al., 2007). The model has five more layers –ends at S4- on top of the C2 features (i.e. S2b, S3, C2b, C3, and S4). The layers are alternating S and C layers. Similarly S layers perform a Gaussian-like operation on their inputs, and C layers perform a max-like operation, which makes the output invariant to small shifts in scale and position. S2b and C2b features model the bypass routes from V2 to TEO (bypassing V4) and from V4 to TE (bypassing TEO) (Nakamura et al., 1993).

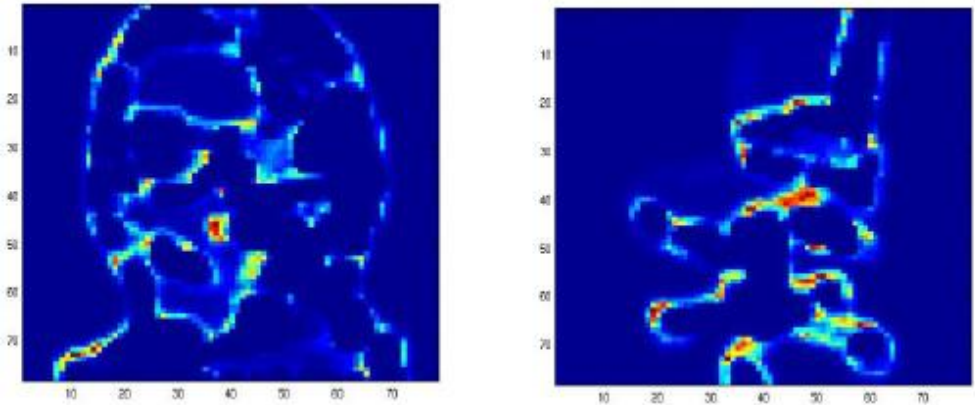

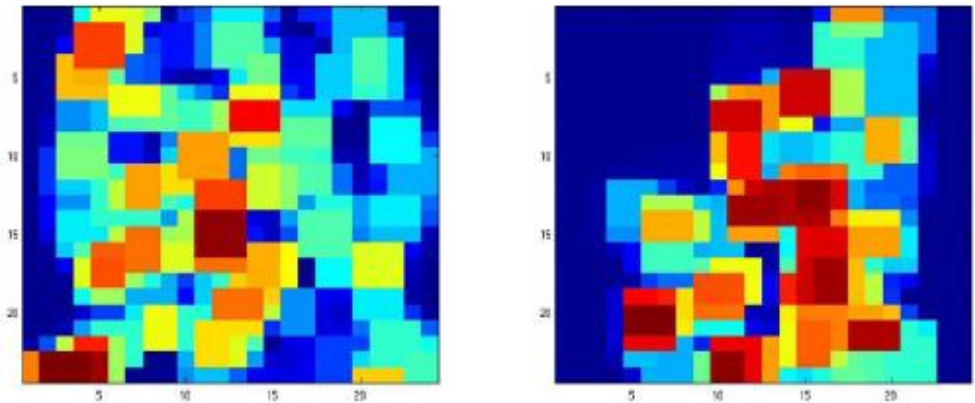

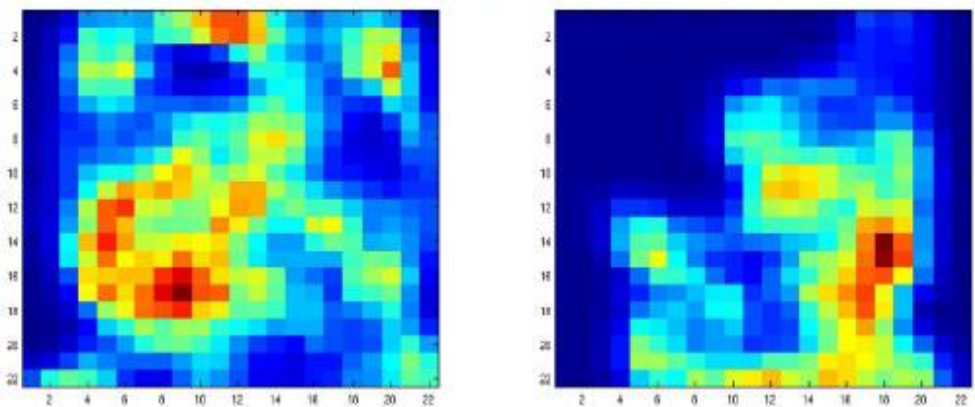

**Figure 22. HMAX output.** The figure shows a sample representation for each of the S2, C2, and S3 layers. The output for C3 layer is a vector of global max values over S3 representations, therefore C3 is not visualized.

In the standard HMAX model, during the training phase, a large number of image patches at various sizes are selected at random. This pool of patches has an important role in the training process and finally in the task of recognizing different objects. Random patch selection is neither efficient nor biologically plausible. In this regard, several studies have been done to select more informative or task-relevant patches. Below I will mention some of these models in chronological order.

Masquelier et al.(2007) used the spike timing-dependent plasticity (STDP) learning rule in an HMAX-like architecture to extract informative visual features. The model is a feedforward spiking neural network with five layers. It operates in the temporal domain and uses spiking neurons. The features extracted by this model have been shown to exhibit robust object recognition in two classification tasks (i.e. face, and motorbike classification). The STDP learning requires each input to be presented several hundred times, whereas our brain is able to learn scenes at a glance. Therefore, the suggested scheme may only account for the learning process of familiar stimuli (e.g. faces).

Meyers and Wolf (2008) built specific features for face identification by a weighted combination of C1 outputs. They obtained the weights using kernelized regularized version of relevant component analysis (KR-RCA) algorithm from a separate set of training images.

Mutch and Lowe (2008) suggested **Sparse Localized Features (SLF)** that is another extension of the standard HMAX model. The model introduces sparsified and localized intermediate-level visual features at the level of C2 layer.

Later, Serre et al. (2010) used an unsupervised clustering algorithm to learn class-specific visual features of intermediate complexity to improve the model performance in a face detection task.

Ghodrati et al. in (2012) proposed the **GMAX model**. In the training phase the model uses feedback from the classification layer (analogous to PFC) to extract informative task-relevant visual features. The model uses an optimization algorithm (i.e. genetic algorithm) to select informative patches from a large pool of patches. Using a subset of training images as validation images and by taking advantage of genetic algorithm a subset of patches that gives the best performance to the classifier are selected. The proposed model uses these optimized features in different object recognition tasks and successfully achieves a high recognition performance.

Another extension was proposed by Rajaei et al. (2012). In which the model uses the adaptive resonance theory (ART) mechanism (Grossberg, 1999) for extracting informative intermediate level visual features. This has made the model stable against forgetting previously learned patterns, which is why the model is called the **Stable Model**. Similar to the HMAX model it extracts C2-like features, except that in the training phase it only selects the highest active C2 units as prototypes that represent the input image. The selection is done using the information coming from top-down connections from C2 layer to C1 layer. The connections match the C1-like features of the input image to the prototypes of the C2 layer. The matching degree is controlled by a vigilance parameter that is fixed separately on a validation set. For a recent comparison between the HMAX-like object recognition models in invariant object recognition see (Ghodrati et al., 2014).

# Conclusion

In this tutorial, I presented a descriptive overview of most of the well-known object-vision models in the literature, and some of the challenges that the models have to deal with. The review started with some basic image representations and mathematical transforms applied to images; then I surveyed state-of-the-art computer vision models and feature extractors; followed by a section on biologically-inspired object-vision models. For most of the models, I provided a visualization of the model output for two sample images drawn from two different categories (animate vs. inanimate). This makes the models readily comparable and provides a very useful and intuitive understanding of the underlying computations in the models. Overall, the review should give the reader a good-level understanding of each of the models, the nature of computations in each, and where they can be applied.

# Acknowledgements

This work was supported by Cambridge Overseas Trust and Yousef Jameel Scholarship to SK.